\documentclass{article}

\usepackage{arxiv}
\usepackage{lineno,hyperref}
\modulolinenumbers[5]
\usepackage{caption}
\usepackage{subcaption}
\usepackage{adjustbox}
\usepackage{graphicx}
\usepackage{amsmath}

\usepackage[utf8]{inputenc} % allow utf-8 input
\usepackage[T1]{fontenc}    % use 8-bit T1 fonts
\usepackage{hyperref}       % hyperlinks
\usepackage{url}            % simple URL typesetting
\usepackage{booktabs}       % professional-quality tables
\usepackage{amsfonts}       % blackboard math symbols
\usepackage{nicefrac}       % compact symbols for 1/2, etc.
\usepackage{microtype}      % microtypography
\usepackage{lipsum}		% Can be removed after putting your text content
\usepackage{graphicx}
\usepackage{natbib}
\usepackage{doi}

\usepackage[table,xcdraw]{xcolor}
\usepackage{graphics}

\usepackage{todonotes}

\title{Tabular Data: Deep Learning is Not All You Need}

 \author{Ravid Shwartz-Ziv
  \\ravid.ziv@intel.com \\
   IT AI Group, Intel
  \And 
  Amitai Armon\\
  amitai.armon@intel.com\\
  IT AI Group, Intel}

\begin{document}

\maketitle


%\author[mymainaddress]{Ravid Shwartz-Ziv\corref{mycorrespondingauthor}}


%\author[mymainaddress,mysecondaryaddress]{IT AI Group, Intel}

%\ead{ravid.ziv@intel.com}

%\author[mysecondaryaddress]{Amitai Armon}
%\address[mymainaddress, mysecondaryaddress]{IT AI Group, Intel}

\begin{abstract}
A key element in solving real-life data science problems is selecting the types of models to use. Tree ensemble models (such as XGBoost) are usually recommended for classification and regression problems with tabular data. However, several deep learning models for tabular data have recently been proposed, claiming to outperform XGBoost for some use cases. This paper explores whether these deep models should be a recommended option for tabular data by rigorously comparing the new deep models to XGBoost on various datasets. In addition to systematically comparing their performance, we consider the tuning and computation they require. Our study shows that XGBoost outperforms these deep models across the datasets, including the datasets used in the papers that proposed the deep models. We also demonstrate that XGBoost requires much less tuning. On the positive side, we show that an ensemble of deep models and XGBoost performs better on these datasets than XGBoost alone. 
\end{abstract}


%\linenumbers
\section{Introduction}
Deep neural networks have demonstrated great success across various domains, including images, audio, and text \citep{devlin2018bert, he2016deep, oord2016wavenet}. There are several canonical architectures for encoding raw data into meaningful representations in these domains. These canonical architectures usually perform well in real-world applications.

In real-world applications, the most common data type is tabular data, comprising samples (rows) with the same set of features (columns). Tabular data is used in practical applications in many fields, including medicine, finance, manufacturing, climate science, and many other applications that are based on relational databases. During the last decade, traditional machine learning methods, such as gradient-boosted decision trees (GBDT) \citep{chen2016xgboost}, still dominated tabular data modeling and showed superior performance over deep learning.  In spite of their theoretical advantages \citep{shwartz2018representation, Poggio30039, piran2020dual}, deep neural networks pose many challenges when applied to tabular data, such as lack of locality, data sparsity (missing values), mixed feature types (numerical, ordinal, and categorical), and lack of prior knowledge about the dataset structure (unlike with text or images). Moreover, deep neural networks are perceived as a "black box" approach -- in other words, they lack transparency or interpretability of how input data are transformed into model outputs \citep{shwartz2017opening}. Although the ``no free lunch'' principle \citep{wolpert1997no} always applies, tree-ensemble algorithms, such as XGBoost, are considered the recommended option for real-life tabular data problems \citep{chen2016xgboost,friedman2001greedy, dorogush2018catboost}.

Recently, there have been several attempts to develop deep networks for tabular data \citep{arik2019tabnet, abutbul2020dnf, popov2019neural}, some of which have been claimed to outperform GBDT. However, each work in this field used different datasets since there is no standard benchmark (such as ImageNet \citep{deng2009imagenet} or GLUE \citep{wang-etal-2018-glue}). As a result, it is challenging to compare tabular data models accurately, noting that some lack open-source implementations. In addition, papers that attempted to compare these models often did not optimize them equally. Thus, it became unclear whether deep learning models for tabular data surpass GBDT and which deep learning models perform best. These obscurities impede the research and development process and make the conclusions from these papers unclear.
As the number of papers on deep learning for tabular data is rising, we believe it is time to rigorously review recent developments in this area and provide more substantiated conclusions that will serve as a basis for future research.  The main purpose of this study is to explore whether any of the recently proposed deep models should indeed be a recommended choice for tabular dataset problems. There are two parts to this question: (1) Are the models more accurate, especially for datasets that did not appear in the paper that proposed them? (2) How long do training and hyperparameter search take in comparison to other models?

To answer these questions, we evaluate these recently proposed deep learning models and XGBoost on diverse tabular datasets with the same tuning protocol.  We analyze the deep models proposed in four recent papers using eleven datasets, nine of which were used in these papers. We show that in most cases, each deep model performs best on the datasets used in its respective paper but significantly worse on other datasets. Additionally, our study shows that XGBoost usually outperforms deep models on these datasets. Furthermore, we demonstrate that the hyperparameter search process was much shorter for XGBoost. On the positive side, we examine the performance of an ensemble of deep models combined with XGBoost and show that this ensemble gives the best results. It also performs better than an ensemble of deep models without XGBoost, or an ensemble of classical models. 

Of course, any selection of tabular datasets cannot represent the full diversity of this type of data, and the ''no free lunch`` principle means that no one model is always better or worse than any other model. Still, our systematic study demonstrates that deep learning is currently not all we need for tabular data, despite the recent significant progress.

The paper is organized as follows: Section \ref{sec:back} provides a background and overview of the models for tabular data. Next, section \ref{sec:results} presents the experimental setup and results. Finally, the discussion, conclusions, and suggested future work are included in section \ref{sec:summary}.
\section{Background}
\label{sec:back}
Traditionally, classical machine learning methods, such as GBDT\citep{chen2016xgboost} dominate tabular data applications due to their superior performance.  Researchers and practitioners use several GBDT algorithms, including XGBoost, LightGBM, and CatBoost \citep{chen2016xgboost, NIPS2017_6449f44a, Prokhorenkova2018}. GBDT learns a series of weak learners to predict the output. In GBDT, the weak learner is the standard decision tree, which lacks differentiability. Despite their differences, their performance on many tasks is similar \citep{Prokhorenkova2018}.

\textbf{XGBoost Model.} The XGBoost algorithm \citep{chen2016xgboost}  is an extendible gradient boosting tree algorithm, that achieves state-of-the-art results on many tabular datasets \citep{zhao2019deep, ramraj2016experimenting}. Gradient boosting is an algorithm in which new models are created from previous models' residuals and then combined to make the final prediction. When adding new models, it uses a gradient descent algorithm to minimize the loss. XGBoost is one of the most popular GBDT implementations.

\subsection{Deep Neural Models for Tabular Data}
\label{sec:tabular_data}

As mentioned above, several recent studies have applied deep learning to the tabular data domain, introducing new neural architectures to achieve improved performance on tabular data \citep{arik2019tabnet, badirli2020gradient, hazimeh2020tree, huang2020tabtransformer, klambauer2017self,baosenguo, popov2019neural}. There are two main categories of these models, which we describe briefly below.

\textbf{Differentiable trees}.  As a result of the excellent performance of ensembles of decision trees on tabular data, this line of work looks for ways to make decision trees differentiable.  Classical decision trees cannot be used as components of end-to-end training pipelines, as they are not differentiable and do not allow gradient optimization. Several works address this issue by smoothing the decision functions in the internal tree nodes to make the tree function and tree routing differentiable \citep{hazimeh2020tree, Kontschieder2015, popov2019neural}.

\textbf{Attention-based models.} Since attention-based models are widely used in different fields, several authors have also proposed using attention-like modules for tabular deep networks. Recent works have suggested both inter-sample attention, in which features of a given sample interact with each other, and intra-sample attention, in which data points interact with each other using entire rows/samples \citep{arik2019tabnet, huang2020tabtransformer, somepalli2021saint}.

In addition, the literature offers other architecture designs that do not fit explicitly into these two categories: (1) Regularization methods that learn a ``regularization strength" for every neural weight, through the use of large-scale hyperparameter tuning schemes \citep{kadra2021regularization, shavitt2018regularization}; (2) Explicit modeling of multiplicative interactions, which considered different ways to incorporate feature products into the MLP model\citep{Beutel2018}; and (3) 1D-CNN, which utilizes the advantages of convolutions in tabular data \citep{baosenguo}.

As noted above, the community developed various models that were evaluated using different benchmarks; these models have rarely been compared.

Among the recently proposed deep models for learning from tabular data, we examine four models that have been claimed to outperform tree ensembles and have attracted significant industry attention: TabNet \citep{arik2019tabnet}, NODE \citep{popov2019neural}, DNF-Net \citep{abutbul2020dnf}, and 1D-CNN \citep{baosenguo}. Below, we briefly describe the key ideas of these models that are relevant to this research.

\textbf{TabNet.} TabNet is a deep learning end-to-end model that performed well across several datasets \citep{arik2019tabnet}. It includes an encoder, in which sequential decision steps encode features using sparse learned masks and select relevant features for each row using the mask (with attention). Using sparsemax layers, the encoder forces the selection of a small set of features. The advantage of learning masks is that feature selection need not be all-or-nothing. Rather than using a hard threshold on a feature, a learnable mask can make a soft decision, thus relaxing classical (non-differentiable) feature selection methods.

\textbf{Neural Oblivious Decision Ensembles (NODE).} The NODE network \citep{popov2019neural} contains equal-depth oblivious decision trees (ODTs), which are differentiable such that error gradients can backpropagate through them.  Like classical decision trees, ODTs split data according to selected features and compare each with a learned threshold. However, only one feature is chosen at each level, resulting in a balanced ODT that can be differentiated. Thus, the complete model provides an ensemble of differentiable trees. 

\textbf{DNF-Net.} The idea behind DNF-Net  \citep{abutbul2020dnf} is to simulate disjunctive normal formulas (DNF) in deep neural networks. The authors proposed replacing the hard Boolean formulas with soft, differentiable versions of them. 
A key feature of this model is the disjunctive normal neural form (DNNF) block, which contains (1) a fully connected layer; and (2) a DNNF layer formed by a soft version of binary conjunctions over literals. The complete model is an ensemble of these DNNFs.  

\textbf{1D-CNN.} Recently, 1D-conventional neural network (CNN) achieved the best single model performance in a Kaggle competition with tabular data \citep{baosenguo}. The model is based on the idea that the CNN structure performs well in feature extraction. Still, it is rarely used in tabular data because the feature ordering has no locality characteristics.  In this model,  an fully connected layer is used to create a larger set of features with locality characteristics, and it is followed by several 1D-Conv layers with shortcut-like connections. 

\subsection{Model Ensemble}
Ensemble learning is a well-known method for improving performance and reducing variance through training multiple models and combining their predictions \citep{caruana2004ensemble}. 
It enhances classifier performance by combining the outputs from many submodels (base learners) trained to solve the same task. The final prediction is obtained by combining the predictions made by each base learner. Consequently, ensembles tend to improve the prediction performance and reduce variance, leading to more stable and accurate results.  Ensemble learning assumes that different machine learning methods may perform better or have mistakes in different situations. Hence, we would expect that a method that uses multiple machine learning methods would produce superior results.
In the literature on ensemble learning, a variety of methods have been explored. Intuitively, using more data for training the base learners helps reduce their bias, and ensembling helps reduce the variance.

Ensemble learning can usually be classified into two main types. The first includes techniques based on randomization, such as random forests \citep{breiman2001random}, where each ensemble member has a different initial parameterization and training data. In this type of models, the base learners can be trained simultaneously without interacting with each other. In the second type of ensembles, boosting-based approaches are used to fit the ensemble base learners sequentially.  In both cases, all the base learners may either use the same architecture or differ in their architecture. Achieving high performance requires the individual base learners to provide sufficiently high performance \citep{breiman2001random}. In most ensembles, decision trees are used as the base learner. 
Breiman \cite{breiman1996bagging}, introduced bagging, a method that combines decision trees generated by randomly selected subsets of the training data and votes on the final outcome. \cite{freund1996experiments} presented the boosting technique, in which the weights of training samples are updated after each iteration of training, and weighted voting is used to combine the classification outputs. In boosting, the new models are added to adjust the existing models' errors. The models are added iteratively until no significant improvements are observed. \cite{wolpert1992stacked} suggested using linear regression to combine the outputs of neural networks, which later became known as stacking.

In our study, we use five classifiers in our ensemble: TabNet, NODE, DNF-Net, 1D-CNN, and XGBoost. To construct a practical and straightforward ensemble, we suggest two different versions: (1) Treating the ensemble as a uniformly weighted mixture model and combining the predictions as 
\begin{align}
    \label{eq:simple_method}
    p(y|x) =\sum_{k=1}^K p_{\theta_m}(y|x, \theta_m)
\end{align}

(2) A weighted average of the predictions from each trained model.  The relative weights of each model are defined simply by the normalized validation loss:
\begin{align}
\label{eq:weighted_method}
 p(y|x) =\sum_{k=1}^K l^\text{val}_k p_{\theta_m}(y|x, \theta_m)   
\end{align}

where $l^\text{val}_k$  is the validation loss for the k-th model. Having uniform weights makes Equation \ref{eq:simple_method} a special case of Equation \ref{eq:weighted_method}. Some of the models above have ensembles built into their design, as mentioned above. Nevertheless, these ensembles are of the same basic submodels with different parameters, not of different types of models.   Since these models typically perform better with more data, we use the entire training dataset to train each model.
\section{Comparing the Models}
\label{sec:results}
We investigate whether the proposed deep models have advantages when used for various tabular datasets. For real-world applications, the models must (1) perform accurately, (2) be trained and make inferences efficiently, and (3) have a short optimization time (fast hyperparameter tuning). Therefore, we first evaluate the performance of the deep models, XGBoost, and ensembles on various datasets. Next, we analyze the different components of the ensemble. Then, we investigate how to select models for the ensemble and test whether deep models are essential for producing good results or combining `classical' models (XGBoost, SVM \citep{cortes1995support} and CatBoost \citep{dorogush2018catboost}) is sufficient. In addition, we explore the tradeoff between accuracy and computational resource requirements. Finally, we compare the hyperparameter search process of the different models and demonstrate that XGBoost outperforms the deep models. 

\subsection{Experimental Setup}
\subsubsection{Data-sets Description}
\label{datasets}

Our experiments use 11 tabular datasets that represent diverse classification and regression problems.  The datasets include 10 to 2,000 features, 1 to 7 classes, and 7,000 to 1,000,000 samples (for a full description, see Table \ref{table:datastes}). Additionally, they differ in the number of numerical and categorical features. In some datasets, you can find ``heterogeneous'' features - which describe the physical properties of an object in different units of measurement. Other datasets contain ``homogeneous'' features, such as pixels for images or words for text. We use nine datasets from the TabNet, DNF-Net, and NODE papers, drawing three datasets from each paper. In addition, we use two Kaggle datasets not used by any of these papers. 1D-CNN was recently proposed in a Kaggle competition for use on one specific dataset, which we do not explore.  Each dataset was preprocessed and trained as described in the original paper. Data is standardized to have zero mean and unit variance, and the statistics for the standardization are calculated based on the training data. The datasets we use are Forest Cover Type, Higgs Boson, and  Year Prediction \citep{Dua2019}, Rossmann Store Sales \citep{rossman2019}, Gas Concentrations, Eye Movements, and Gesture Phase \citep{vanschoren2014openml}, MSLR \citep{QinL13}, Epsilon \citep{epsilon}, Shrutime \citep{churn}, and Blastchar \citep{blastchar2019}.

 \begin{table}[]
 \resizebox{\columnwidth}{!}{%
\begin{tabular}{lllllll}
\rowcolor[HTML]{82b8e9} 
\multicolumn{1}{l}{\cellcolor[HTML]{82b8e9}\textbf{Dataset}}                & \multicolumn{1}{l}{\cellcolor[HTML]{82b8e9}\textbf{Features}} & \multicolumn{1}{l}{\cellcolor[HTML]{82b8e9}\textbf{Classes}} & \multicolumn{1}{l}{\cellcolor[HTML]{82b8e9}\textbf{Samples}} & \multicolumn{1}{l}{\cellcolor[HTML]{82b8e9}\textbf{Source}}                  & \multicolumn{1}{l}{\cellcolor[HTML]{82b8e9}\textbf{Paper}}                                                                \\
\noalign{\global\arrayrulewidth=0.3mm} \arrayrulecolor[HTML]{000000}\hline
\noalign{\global\arrayrulewidth=0.6mm} \arrayrulecolor[HTML]{FFFFFF}\hline

\rowcolor[HTML]{D2DEEF} 
Gesture   Phase        & 32       & 5       & 9.8k    & OpenML                  & DNF-Net                                                \\
\rowcolor[HTML]{EAEFF7} 
Gas   Concentrations   & 129      & 6       & 13.9k   & OpenML                  & DNF-Net                                                       \\
\rowcolor[HTML]{D2DEEF} 
Eye   Movements        & 26       & 3       & 10.9k   & OpenML                  & DNF-Net                                        \\
\rowcolor[HTML]{EAEFF7} 
Epsilon                & 2000     & 2       & 500k    & PASCAL   Challenge 2008 & NODE        \\
\rowcolor[HTML]{D2DEEF} 
YearPrediction         & 90       & 1       & 515k    & Million   Song Dataset  & NODE                   \\
\rowcolor[HTML]{EAEFF7} 
Microsoft (MSLR)             & 136      & 5       & 964k    & MSLR-WEB10K             & NODE                        \\
\rowcolor[HTML]{D2DEEF} 
Rossmann   Store Sales & 10       & 1       & 1018K   & Kaggle                  & TabNet                             \\
\rowcolor[HTML]{EAEFF7} 
Forest   Cover Type    & 54       & 7       & 580k    & Kaggle                  & TabNet                        \\
\rowcolor[HTML]{D2DEEF} 
Higgs   Boson         & 30       & 2       & 800k    & Kaggle                  & TabNet                                 \\
\rowcolor[HTML]{EAEFF7} 
Shrutime               & 11       & 2       & 10k     & Kaggle                  & New   dataset                \\
\rowcolor[HTML]{D2DEEF} 
Blastchar              & 20       & 2       & 7k      & Kaggle                  & New   dataset              
\end{tabular}%
}
\caption{Description of the tabular datasets}
\label{table:datastes}
\end{table}

\subsubsection{Implementation Details}
\textbf{The Optimization Process}. To select the model hyperparameters, we used HyperOpt \citep{bergstra2015hyperopt}, which uses Bayesian optimization. The hyperparameter search was run for $1,000$ steps on each dataset by optimizing the results on a validation set. The initial hyperparameters were taken from the original paper. Each model had 6-9 main hyperparameters that we optimized.  For the deep learning model, these include the learning rate, number of layers, and number of nodes. The full hyperparameter search space for each model is provided in \ref{app:hyperparameters}.

We split the datasets into training, validation, and test sets in the same way as in the original papers that used them. When the split was reported to be random, we performed three repetitions of the random partition (as done in the original paper), and we reported their mean and the standard error of the mean.  Otherwise, we used four random seed initializations in the same partition, and we report their average.

\textbf{Metrics and evaluation.} For binary classification problems, we report the cross-entropy loss. For regression problems, we report the root mean square error.  We ran four experiments with different random seeds for each tuned configuration, and we reported the performance on the test set. 

\textbf{Statistical significance test}.
In addition to using the RMSE or cross-entropy loss for evaluating the performance of the models, it is also necessary to assess whether these differences are statistically significant. Friedman's test \citep{FriedmanTest} is a widely used nonparametric method for testing statistical significance. Its advantage is that it does not assume that the data in each group is normally distributed. Using the Friedman test, we compare the different models' errors to see whether there is a statistically significant difference between them. Friedman's hypothesis states that the performance results come from the same population.
To test this hypothesis, we examine whether the rank sums of the k classifiers included in the test differ significantly. After applying the omnibus Friedman test, we compare all classifiers against each other or against a baseline classifier. The significance level (95\% in our study) determines whether the null hypothesis should be rejected based on the resulting p-value for each model pair. If the p-value is greater than 0.05, it will fail to reject the null hypothesis for this pair. Otherwise, if the p-value is less than 0.05, the null hypothesis will be rejected at a confidence level of 95\%.

\textbf{Training.} For classification datasets, we minimize cross-entropy loss, while for regression datasets, we minimize and report mean squared error. We use the term ``original model'' to refer to the model used on a given dataset in the paper that presented the respective model. The ``unseen datasets'' for each model are those not mentioned in the paper that published the respective model. Note that a model's unseen dataset is not a dataset it was not trained on, but a dataset that did not appear in its original paper. For the deep models, we follow the original implementations and use the Adam optimizer \citep{kingma2014adam} without learning rate schedules. For each dataset, we also optimize the batch size for all models. We continue training until there are $100$
consecutive epochs without improvement on the validation set.

\subsection{Results}

\subsubsection*{Do the deep models generalize well to other datasets?}
We first explore whether the deep models perform well when trained on datasets that were not included in their original paper and compare them to XGBoost. 
Table \ref{table:results}  presents the mean and the standard error of the mean of the performance measure for each model for each dataset (a lower value indicates better performance). Columns 1-3 correspond to datasets from the TabNet paper, columns 4–6 to the DNF-Net paper, and columns 7–9 to the NODE paper. The last two columns correspond to datasets that did not appear in any of these papers. 

As mentioned above, the Friedman test, with a 95\% significance level, was carried out to check statistically significant differences between susceptibility models.

We make several observations regarding these results:  
\begin{itemize}
\item In most cases, the models perform worse on unseen datasets than do the datasets' original models.
\item The XGBoost model generally outperformed the deep models. For 8 of the 11 datasets, XGBoost outperformed the deep models, which did not appear in the original paper. For these datasets, the results were significant ($p<0.005$).

\item No deep model consistently outperformed the others. Each deep model was better only on the datasets that appeared in its own paper. However, the performance of the 1D-CNN model may seem better since all datasets were new to it.
 \item The ensemble of deep models and XGBoost outperformed the other models in most cases. For 7 of the 11 datasets, the ensemble of deep models or XGBoost was significantly better than the single deep models.  The p-value in these cases was less than $0.005$, which indicates the null hypothesis (i.e., no difference between the performance of the tested models) is rejected.
\end{itemize}

To directly compare between the different models, we calculate for each dataset the relative performance of each model compared to the best model for that dataset. Table \ref{table:compare} presents the averaged relative performance per model on all its unseen datasets (geometric mean). The ensemble of all models was the best model with $2.32\%$ average relative increase (all p-values except one were less than 0.005). XGBoost was the second best with $3.4\%$, 1D-CNN had $7.5\%$, TabNet had $10.5\%$, DNF-Net had $11.8\%$, and NODE had $14.2\%$.

These results are  surprising. When we trained on datasets other than those in their original papers, the deep models performed worse than XGBoost. Compared to XGBoost and the full ensemble, the single deep model's performance is much more sensitive to the specific dataset. There may be several reasons for the deep models' lower performance when they are trained on previously unseen datasets. The first possibility is \textbf{selection bias}. Each paper may have naturally demonstrated the model's performance on datasets with which the model worked well. The second possibility is differences in the \textbf{optimization of hyperparameters}. Each paper may have set the model's hyperparameters based on a more extensive hyperparameter search on the datasets presented in that paper, resulting in better performance. Our results for each model on its original datasets matched those presented in its respective paper, thus excluding implementation issues as the possible reason for our observations.

\begin{center} 
 \begin{table}[]
% \scalebox{0.8}{

\resizebox{\columnwidth}{!}{%
\begin{tabular}{lrrrrrrr}
\rowcolor[HTML]{82b8e9} 
{\color[HTML]{000000} \textbf{Model Name}} &
\multicolumn{1}{l}{\cellcolor[HTML]{82b8e9}{\color[HTML]{000000} \textbf{Rossman}}} & \multicolumn{1}{l}{\cellcolor[HTML]{82b8e9}{\color[HTML]{000000}\textbf{ CoverType}}} & \multicolumn{1}{l}{\cellcolor[HTML]{82b8e9}{\color[HTML]{000000} \textbf{Higgs}}} & \multicolumn{1}{l}{\cellcolor[HTML]{82b8e9}{\color[HTML]{000000} \textbf{Gas}}} & \multicolumn{1}{l}{\cellcolor[HTML]{82b8e9}{\color[HTML]{000000}\textbf{ Eye}}} & \multicolumn{1}{l}{\cellcolor[HTML]{82b8e9}{\color[HTML]{000000} \textbf{Gesture}}}   \\
\noalign{\global\arrayrulewidth=0.3mm} \arrayrulecolor[HTML]{000000}\hline
\noalign{\global\arrayrulewidth=0.6mm} \arrayrulecolor[HTML]{FFFFFF}\hline

\rowcolor[HTML]{D2DEEF} 
XGBoost & $490.18 \pm 1.19$ & $3.13\pm 0.09$ & $21.62\pm 0.33 $ & $2.18 \pm 0.20$ &\textbf{56.07}$\pm 0.65$ & $80.64 \pm 0.80$ \\
\noalign{\global\arrayrulewidth=0.3mm} \arrayrulecolor[HTML]{000000}\hline
\rowcolor[HTML]{EAEFF7} 
NODE & $488.59 \pm 1.24$      &$4.15 \pm 0.13$&  $21.19 \pm 0.69$ & $2.17 \pm 0.18$ & $68.35 \pm 0.66$ & $92.12 \pm 0.82$ \\
\rowcolor[HTML]{D2DEEF} 
DNF-Net & $503.83 \pm 1.41$ & $3.96 \pm 0.11$ & $23.68 \pm 0.83$ & \textbf{1.44} $\pm 0.09$&  $68.38 \pm 0.65$ & $86.98 \pm 0.74$\\
\rowcolor[HTML]{EAEFF7} 
TabNet & \textbf{485.12}$\pm 1.93$ &  $3.01 \pm 0.08$ & \textbf{21.14}$\pm 0.20$&  $1.92 \pm 0.14$ & $67.13 \pm 0.69$ & $96.42 \pm 0.87$\\
\rowcolor[HTML]{D2DEEF} 
1D-CNN & $493.81 \pm 2.23 $ & $3.51 \pm 0.13$ & $22.33 \pm 0.73$ & $1.79\pm 0.19$ & $67.9 \pm 0.64$ &$97.89 \pm 0.82$  \\
\noalign{\global\arrayrulewidth=0.3mm} \arrayrulecolor[HTML]{000000}\hline
\rowcolor[HTML]{EAEFF7} 
Simple Ensemble & $488.57 \pm 2.14$ & $3.19 \pm 0.18$ & $22.46 \pm 0.38$ & $2.36\pm 0.13$ & $58.72 \pm 0.67$ & $89.45 \pm 0.89$ \\
\rowcolor[HTML]{D2DEEF} 
Deep Ensemble w/o XGBoost & $489.94 \pm 2.09$ & $3.52\pm 0.10$ & $22.41 \pm 0.54$ & $1.98 \pm 0.13$ & $69.28 \pm 0.62$ & $93.50 \pm 0.75$\\
\rowcolor[HTML]{EAEFF7} 
Deep Ensemble w XGBoost & $485.33 \pm 1.29$ & \textbf{2.99} $\pm 0.08$& $22.34 \pm  0.81$ & $1.69 \pm 0.10$ & $59.43 \pm 0.60$& \textbf{78.93} $\pm 0.73$ \\

 \multicolumn{1}{c}{} & \multicolumn{3}{c}{\raisebox{.5\normalbaselineskip}[0pt][0pt]{$\underbrace{\phantom{abcdefgh}\hspace*{20\tabcolsep}}$}} & \multicolumn{3}{c}{\raisebox{.5\normalbaselineskip}[0pt][0pt]{$\underbrace{\phantom{abcdefgh}\hspace*{20\tabcolsep}}$}} \\[1ex] 
 \multicolumn{1}{c}{} & \multicolumn{3}{c}{TabNet}& \multicolumn{3}{c}{DNF-Net}
\end{tabular}%
}
\resizebox{\columnwidth}{!}{%
\begin{tabular}{lrrrrr}
\\
\rowcolor[HTML]{82b8e9} 
{\color[HTML]{000000} \textbf{Model Name}} &
 \multicolumn{1}{l}{\cellcolor[HTML]{82b8e9}{\color[HTML]{000000} \textbf{YearPrediction}}} & \multicolumn{1}{l}{\cellcolor[HTML]{82b8e9}{\color[HTML]{000000}\textbf{ MSLR}}} & \multicolumn{1}{l}{\cellcolor[HTML]{82b8e9}{\color[HTML]{000000} \textbf{Epsilon}}} & \multicolumn{1}{l}{\cellcolor[HTML]{82b8e9}{\color[HTML]{000000}\textbf{ Shrutime}}} & \multicolumn{1}{l}{\cellcolor[HTML]{82b8e9}{\color[HTML]{000000}\textbf{ Blastchar}}} \\
\noalign{\global\arrayrulewidth=0.3mm} \arrayrulecolor[HTML]{000000}\hline
\noalign{\global\arrayrulewidth=0.6mm} \arrayrulecolor[HTML]{FFFFFF}\hline

\rowcolor[HTML]{D2DEEF} 
XGBoost & $77.98 \pm 0.11$ & $55.43 \pm $2e-2& $11.12 \pm $3e-2 & $13.82 \pm 0.19$ &$20.39\pm 0.21$ \\
\noalign{\global\arrayrulewidth=0.3mm} \arrayrulecolor[HTML]{000000}\hline

\rowcolor[HTML]{EAEFF7} 
NODE & $76.39 \pm0.13$ &$55.72 \pm $3e-2 & \textbf{10.39}$\pm$1e-2&  $14.61 \pm 0.10$ & $21.40\pm 0.25$ \\
\rowcolor[HTML]{D2DEEF} 
DNF-Net &  $81.21 \pm 0.18$ & $56.83 \pm $3e-2 & $12.23 \pm$4e-2 & $16.8\pm 0.09$ &$27.91\pm0.17$ \\
\rowcolor[HTML]{EAEFF7} 
TabNet & $83.19 \pm 0.19$ & $56.04 \pm$1e-2 & $11.92 \pm$3e-2  & $14.94\pm, 0.13$ & $23.72\pm0.19$ \\
\rowcolor[HTML]{D2DEEF} 
1D-CNN &  $78.94 \pm 0.14$ & $55.97 \pm $4e-2 & $11.08 \pm$6e-2 & $15.31\pm 0.16$ & $24.68\pm0.22$ \\
\noalign{\global\arrayrulewidth=0.3mm} \arrayrulecolor[HTML]{000000}\hline
\rowcolor[HTML]{EAEFF7} 
Simple Ensemble & $78.01\pm 0.17$ & $55.46 \pm $4e-2 & $11.07\pm$4e-2 & $13.61\pm, 0.14$ & $21.18\pm 0.17$ \\
\rowcolor[HTML]{D2DEEF} 
Deep Ensemble w/o XGBoost &  $78.99 \pm 0.11$ & $55.59\pm $3e-2 & $10.95 \pm$1e-2 & $14.69 \pm 0.11$ & $24.25\pm 0.22$ \\
\rowcolor[HTML]{EAEFF7} 
Deep Ensemble w XGBoost & \textbf{76.19} $\pm 0.21$&  \textbf{55.38}$\pm$1e-2 & $11.18 \pm$1e-2 &\textbf{13.10}$\pm 0.15$& \textbf{20.18}$\pm 0.16$ \\

 \multicolumn{1}{c}{} &  \multicolumn{3}{c}{\raisebox{.5\normalbaselineskip}[0pt][0pt]{$\underbrace{\phantom{abcdefgh}\hspace*{26\tabcolsep}}$}}& \multicolumn{2}{c}{\raisebox{.5\normalbaselineskip}[0pt][0pt]{$\underbrace{\phantom{abcdefgh}\hspace*{14\tabcolsep}}$}}\\[1ex] 
 \multicolumn{1}{c}{} & \multicolumn{3}{c}{NODE} &\multicolumn{2}{c}{New datasets}
\end{tabular}%
}
\caption{\textbf{Test results on tabular datasets.} Presenting the performance for each model. MSE is presented for the YearPrediction and Rossman datasets, and cross-entropy loss (with $100$X factor) is presented for the other datasets. The papers that used these datasets are indicated below the table. The values are the averages of four training runs (lower value is better), along with the standard error of the mean (SEM)}
\label{table:results}
\end{table}

\end{center}

\begin{table}[]
\begin{center}
\scalebox{0.8}{

%\begin{adjustbox}{width=0.5\columnwidth,center}
\begin{tabular}{lr}
\rowcolor[HTML]{82b8e9} 
\multicolumn{1}{l}{\cellcolor[HTML]{82b8e9}\textbf{Name}} & {\cellcolor[HTML]{82b8e9}\begin{tabular}[c]{@{}l@{}} \textbf{Average Relative} \\ \textbf{ Performance} (\%)\end{tabular} } \\
\rowcolor[HTML]{D2DEEF} 
XGBoost &  3.34 \\
\rowcolor[HTML]{EAEFF7} 
NODE & 14.21 \\
\rowcolor[HTML]{D2DEEF} 
DNF-Net &  11.96 \\
\rowcolor[HTML]{EAEFF7} 
TabNet & 10.51 \\
\rowcolor[HTML]{D2DEEF} 
1D-CNN & 7.56 \\
\rowcolor[HTML]{EAEFF7} 
Simple Ensemble & 3.15 \\
\rowcolor[HTML]{D2DEEF} 
Deep Ensemble w/o XGBoost & 6.91 \\
\rowcolor[HTML]{EAEFF7} 
Deep Ensemble w XGBoost & \textbf{2.32} \\
\end{tabular}}
\end{center}
\caption{Average relative performance deterioration for each model on its unseen datasets (lower value is better).}
\label{table:compare}
%\end{adjustbox}
\end{table}

\iffalse % covered by the previous subsection
\subsubsection*{Performance on new datasets}
Following our observation that the deep models did not perform well on `unseen datasets', we tested their performance on two new data sets from Kaggle competition: the shrutime dataset and the Blastchar dataset.
Table \ref{table:results} shows the results of the models on these two datasets. XGBoost performs better than individual deep networks on both datasets, but the ensemble of all models provides the best results.
\fi

\subsubsection*{Do we need both XGBoost and deep networks?}

In the previous subsection, we saw that the ensemble of XGBoost and deep models performed best across the datasets. It is therefore interesting to examine which component of our ensemble is mandatory. One question is whether XGBoost needs to be combined with the deep models, or would a simpler ensemble of nondeep models perform similarly. To explore this, we trained an ensemble of XGBoost and other nondeep models: SVM \citep{cortes1995support} and CatBoost\citep{dorogush2018catboost}.
Table \ref{table:results} shows that the ensemble of classical models performed much worse than the ensemble of deep networks and XGBoost. Additionally, the table shows that the ensemble of deep models alone (without XGBoost) did not provide good results. These differences were significant (all p-values were extremely low and less than $0.005$). This indicates that combining both the deep models and XGBoost provides an advantage for these datasets.

\subsubsection*{Subset of models}
We observed that the ensemble improved performance, but the use of multiple models also requires additional computation. When real-world applications are considered, computational constraints may affect the eventual model performance. We therefore considered using subsets of models within the ensemble to examine the tradeoff between performance and computation.

There are several ways to choose a subset from an ensemble of models: (1) based on the \textbf{validation loss}, choosing models with low validation loss first; (2) based on the models' \textbf{uncertainty for each example}, choosing the highest confidence models (by some uncertainty measure) for each example; and (3) based on a \textbf{random order}.

\begin{figure}[ht]
\centering
 \centering
 \includegraphics[width=\linewidth]{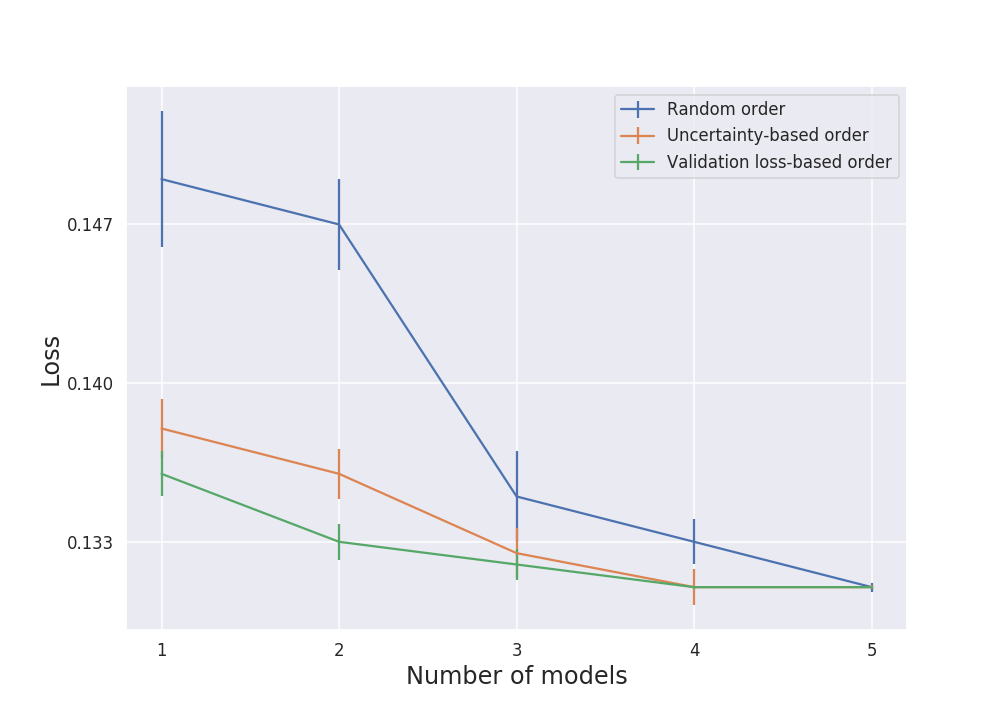}
 \captionof{figure}{\textbf{The impact of selecting a \\ subset of models in the ensemble}.}
 \label{fig:subset}
\end{figure}

In Figure \ref{fig:subset}, these methods of selecting models are compared for an example of an unseen dataset (Shrutime). The best selection approach was averaging the predictions based on the models' validation loss. Only three models were needed to achieve almost optimal performance this way. On the other hand, choosing the models randomly provided the worst choice according to our comparison. The differences in performance for the first three models were significant ($p<0.005$).

\subsubsection*{How difficult is the optimization?}

In real-life settings, there are limited time and resources to train a model for a new dataset and optimize its hyperparameters. It is therefore interesting to understand how difficult it is to do this for each model. One approach to evaluating this is to calculate the number of computations the model requires. This is typically measured in floating-point operations per second (FLOPS). Tang et al. \cite{Tang2018experimental} suggested that the FLOPS number is a better indicator of energy usage and latency than the number of parameters. However, each hyperparameter set has a different FLOPS number, so it is impossible to compare different models this way when optimizing the hyperparameters. Another approach is to compare the total time it takes to train and optimize the model.  Generally, we found XGBoost to be significantly faster than the deep networks in our experiments (more than an order of magnitude).  However, runtime differences are significantly affected by the level of optimization of the software. Comparing a widely used library with non-optimized implementations from papers would not be fair.  Thus, we used another approach: comparing the number of iterations of the hyperparameter optimization process until we reached a plateau. This can be seen as a proxy measure for the model's robustness and how far the default hyperparameters are from the best solutions. It represents an inherent characteristic of the model that does not depend on software optimization.

\begin{figure}[ht]
 \centering
 \includegraphics[width=\linewidth]{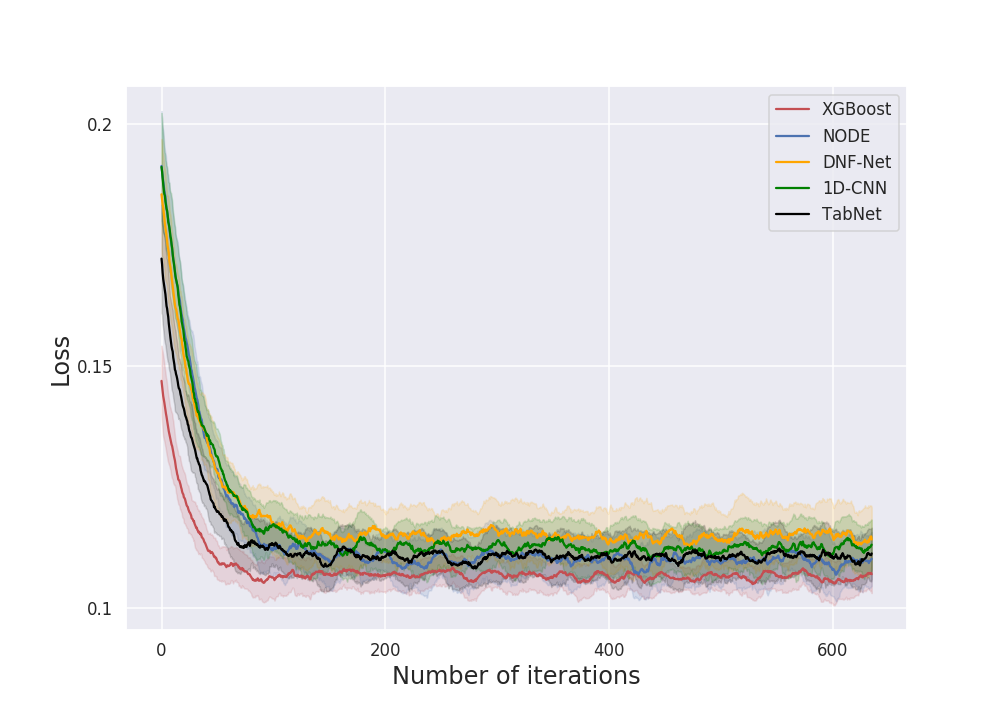}
 \captionof{figure}{\textbf{The Hyper-parameters optimization process for different models}. }
 \label{fig:opt_time}
\end{figure}

Figure \ref{fig:opt_time} shows the model's performance (mean and standard error of the mean) as a function of the number of iterations of the hyperparameter optimization process for the Shrutime dataset. We observe that XGBoost outperformed the deep models, converging to good performance more quickly (in fewer iterations, which were also shorter in terms of runtime). These results may be affected by several factors: (1) We used a \textbf{Bayesian hyperparameter optimization process}, and the results may differ for other optimization processes; (2) \textbf{The initial hyperparameters} of XGBoost may be more robust because they might have been originally suggested based on more datasets. Perhaps we could find some initial hyperparameters that would work better for the deep models for different datasets; and (3) The XGBoost model may have some \textbf{inherent characteristics} that make it more robust and easier to optimize. It may be interesting to investigate this behavior further.

\section{Discussion and Conclusions}
\label{sec:summary}
In this study, we investigated the performance of recently proposed deep models for tabular datasets. In our analysis, the deep models were weaker on datasets that did not appear in their original papers, and they were weaker than XGBoost, the baseline model. Therefore, we proposed using an ensemble of these deep models with XGBoost. This ensemble performed better on these datasets than any individual model and the `non-deep' classical ensemble. In addition, we explored possible tradeoffs between performance, computational inference cost, and hyperparameter optimization time, which are important in real-world applications. Our analysis shows that we must take the reported deep models' performance with a grain of salt. When we made a fair comparison of their performance on other datasets, they provided weaker results. Additionally, it is much more challenging to optimize deep models than XGBoost on a new dataset. However, we found that an ensemble of XGBoost with deep models yielded the best results for the datasets we explored. 

Consequently, when researchers choose a model to use in real-life applications, they must consider several factors.  Apparently, under time constraints, XGBoost may achieve the best results and be the easiest to optimize. However,  XGBoost alone may not be enough to achieve the best performance; we may need to add deep models to our ensemble to maximize performance. 

In conclusion, despite significant progress using deep models for tabular data, they do not outperform XGBoost on the datasets we explored, and further research is probably needed in this area. Taking our findings into account, future research on tabular data must systematically check the performance on several diverse datasets. Our improved ensemble results provide another potential avenue for further research. Nevertheless, this is not sufficient. Comparing models requires including details about how easy it is to identify the most appropriate hyperparameters. Our study showed that some deep models require significantly more attempts to find the correct configuration. Another line of research could be developing new deep models that are easier to optimize and may compete with XGBoost in terms of this parameter.

\bibliographystyle{unsrtnat}
\bibliography{main_q}
\appendix
\section{Tabular Data-sets Description}
\label{app:datasets}
We use a wide variety of datasets, each with different characteristics, such as the number of features, the number of classes, and the number of samples. The datasets include both classification and regression tasks, including in high-dimensional space. In some datasets, you can find "heterogeneous" features - which describe the physical properties of an object in different units of measurement. Other datasets contain homogeneous features, such as pixels for images or words for text. We followed the training and prepossessing procedures outlined in the original study for each dataset. The datasets we use are Forest Cover Type, Higgs Boson and  Year Prediction datasets \citep{Dua2019}, Rossmann Store Sales \citep{rossman2019}, Gas Concentrations, Eye Movements and Gesture Phase \citep{vanschoren2014openml}, MSLR \citep{QinL13}, Epsilon \citep{epsilon}, Shrutime \citep{churn}, and Blastchar \citep{blastchar2019}. (see Table \ref{table:datastes}).
 \begin{table}[]
 \resizebox{\columnwidth}{!}{%
\begin{tabular}{lllllll}
\rowcolor[HTML]{82b8e9} 
\multicolumn{1}{l}{\cellcolor[HTML]{82b8e9}\textbf{Dataset}}                & \multicolumn{1}{l}{\cellcolor[HTML]{82b8e9}\textbf{Features}} & \multicolumn{1}{l}{\cellcolor[HTML]{82b8e9}\textbf{Classes}} & \multicolumn{1}{l}{\cellcolor[HTML]{82b8e9}\textbf{Samples}} & \multicolumn{1}{l}{\cellcolor[HTML]{82b8e9}\textbf{Source}}                  & \multicolumn{1}{l}{\cellcolor[HTML]{82b8e9}\textbf{Paper}}         & \multicolumn{1}{l}{\cellcolor[HTML]{82b8e9}\textbf{Link}}                                                                 \\
\noalign{\global\arrayrulewidth=0.3mm} \arrayrulecolor[HTML]{000000}\hline
\noalign{\global\arrayrulewidth=0.6mm} \arrayrulecolor[HTML]{FFFFFF}\hline

\rowcolor[HTML]{D2DEEF} 
Gesture   Phase        & 32       & 5       & 9.8k    & OpenML                  & DNF-Net       & openml.org/d/4538                                                     \\
\rowcolor[HTML]{EAEFF7} 
Gas   Concentrations   & 129      & 6       & 13.9k   & OpenML                  & DNF-Net       & openml.org/d/1477                                                     \\
\rowcolor[HTML]{D2DEEF} 
Eye   Movements        & 26       & 3       & 10.9k   & OpenML                  & DNF-Net       & openml.org/d/1044                                                     \\
\rowcolor[HTML]{EAEFF7} 
Epsilon                & 2000     & 2       & 500k    & PASCAL   Challenge 2008 & NODE          & https://www.csie.ntu.edu.tw/~cjlin/libsvmtools/datasets/binary.html \\
\rowcolor[HTML]{D2DEEF} 
YearPrediction         & 90       & 1       & 515k    & Million   Song Dataset  & NODE          & https://archive.ics.uci.edu/ml/datasets/yearpredictionmsd             \\
\rowcolor[HTML]{EAEFF7} 
Microsoft (MSLR)             & 136      & 5       & 964k    & MSLR-WEB10K             & NODE          & https://www.microsoft.com/en-us/research/project/mslr/                \\
\rowcolor[HTML]{D2DEEF} 
Rossmann   Store Sales & 10       & 1       & 1018K   & Kaggle                  & TabNet        & https://www.kaggle.com/c/rossmann-store-sales                         \\
\rowcolor[HTML]{EAEFF7} 
Forest   Cover Type    & 54       & 7       & 580k    & Kaggle                  & TabNet        & https://www.kaggle.com/c/forest-cover-type-prediction                 \\
\rowcolor[HTML]{D2DEEF} 
Higges   Boson         & 30       & 2       & 800k    & Kaggle                  & TabNet        & https://www.kaggle.com/c/higgs-boson                                  \\
\rowcolor[HTML]{EAEFF7} 
Shrutime               & 11       & 2       & 10k     & Kaggle                  & New   dataset & https://www.kaggle.com/shrutimechlearn/churn-modelling                \\
\rowcolor[HTML]{D2DEEF} 
Blastchar              & 20       & 2       & 7k      & Kaggle                  & New   dataset & https://www.kaggle.com/blastchar/telco-customer-churn                
\end{tabular}%
}
\caption{Description of the tabular datasets}
\label{appen:table:datastes}
\end{table}
\section{Optimization of hyperparameters}
\label{app:hyperparameters}
To tune the hyperparameters, we split the datasets into training, validation, and test sets in the same way as in the original papers that used them.  Specifically, we performed a random stratified split of the full training data into
train set (80\%) and validation set (20\%) for the Epsilon, YearPrediction, MSLR, Shrutime, and Blastchar datasets. For Eye, Gesture, and Year datasets, we split the full data into validation set (10\%), test set (20\%), and train set (70\%). For Forest Cover Type,  we use the train/val/test split provided by the dataset authors \citep{mitchell2018xgboost}. For the Rossmann dataset, we used the same preprocessing and data split as \citep{dorogush2018catboost} – data from 2014 was used for training and validation, whereas 2015 was used for testing. We split $100k$ samples for validation from the training dataset, and after the optimization of the hyperparameters, we retrained on the entire training dataset. For the Higgs dataset,  we split $500k$ samples for validation from the training dataset, and after the optimization of the hyperparameters, we retrained on the entire training dataset. 
We used the Hyperopt library to optimize the models. We selected the set of hyperparameters corresponding to the smallest loss on the validation set for the final configuration. For all models, early stopping is applied using the validation set.
\subsection{CATBOOST}
The list of hyperparameters and their search spaces for Catboost:
\begin{itemize}
    \item Learning rate: Log-Uniform distribution $[e^{-5}, 1]$
    \item Random strength: Discrete uniform distribution $[1, 20]$
    \item Max size: Discrete uniform distribution $[0, 25]$
    \item L2 leaf regularization: Log-Uniform distribution $[1, 10]$
    \item Bagging temperature: Uniform distribution $[0, 1]$
    \item Leaf estimation iterations: Discrete uniform distribution $[1, 20]$
\end{itemize}
\subsection{XGBoost}
The list of hyperparameters and their search spaces for XGBoost:
\begin{itemize}
    \item Number of estimators: Uniform distribution $[100, 4000]$
    \item Eta: Log-Uniform distribution $[e^{-7}, 1]$
     \item Max depth: Discrete uniform distribution $[1, 10]$
     \item Subsample: Uniform distribution $[0.2, 1]$
     \item Colsample bytree: Uniform distribution $[0.2, 1]$
     \item Colsample bylevel: Uniform distribution $[0.2, 1]$
     \item Min child weight: Log-Uniform distribution $[e^{-16}, e^5]$
     \item Alpha: Uniform choice \{0, Log-Uniform distribution $[e^{-16}, e^2]$\}
     \item Lambda: Uniform choice \{0, Log-Uniform distribution $[e^{-16}, e^2]$\}
     \item Gamma: Uniform choice \{0, Log-Uniform distribution $[e^{-16}, e^2]$\}
\end{itemize}
\subsection{NODE}
The list of hyperparameters and their search spaces for NODE: 
\begin{itemize}
    \item Learning rate: Log-Uniform distribution $[e^{-5}, 1]$
    \item Num layers: Discrete uniform distribution $[1, 10]$
    \item Total tree count: \{$256, 512, 1024, 2048$\}
    \item Tree depth: Discrete uniform distribution $[4, 9]$
    \item Tree output dim: Discrete uniform distribution $[1, 5]$
    \item Learning rate - Log-Uniform distribution $[e^{-4}, 0.5]$
    \item Batch size - Uniform choice \{512, 1024, 2048, 4096, 8192\}

\end{itemize}
\subsection{TabNet}
The list of hyperparameters and their search spaces for TabNet:
\begin{itemize}
    \item Learning rate: Log-Uniform distribution $[e^{-5}, 1]$
    \item feature dim: Discrete uniform distribution $[20, 60]$
    \item output dim: Discrete uniform distribution $[20, 60]$
    \item n steps: Discrete uniform distribution $[1, 8]$
    \item bn epsilon: Uniform distribution $[e^{-5}, e^{-1}]$
    \item relaxation factor: Uniform distribution $[0.3, 2]$
     \item Batch size - Uniform choice \{512, 1024, 2048, 4096, 8192\}

\end{itemize}
\subsection{DNF-Net}
The list of hyperparameters and their search spaces for DNF-Net:
\begin{itemize}
    \item n. formulas: Discrete uniform distribution $[256, 2048]$
    \item Feature selection beta: Discrete uniform distribution $[1e^{-2}, 2]$
   \item Learning rate - Log-Uniform distribution $[e^{-4}, 0.5]$
   \item Batch size - Uniform choice \{512, 1024, 2048, 4096, 8192\}
\end{itemize}
\subsection{1D-CNN}
The list of hyperparameters and their search spaces for 1D-CNN:
\begin{itemize}
    \item Hidden layer sizes: Discrete uniform distribution $[100, 4000]$
    \item Number of layers:Discrete uniform distribution $[1, 6]$
   \item Learning rate - Log-Uniform distribution $[e^{-4}, 0.5]$
   \item Batch size - Uniform choice \{512, 1024, 2048, 4096, 8192\}
\end{itemize}
\end{document}